\documentclass[preprint,12pt,authoryear]{elsarticle}
\journal{Medical Image Analysis}

\makeatletter
\def\ps@pprintTitle{%
 \let\@oddhead\@empty
 \let\@evenhead\@empty
 \def\@oddfoot{}%
 \let\@evenfoot\@oddfoot}
\makeatother

\usepackage{setspace}
\usepackage{graphicx,natbib}
\usepackage{float, subcaption}
\usepackage{amssymb,amsmath,multicol, multirow}
\usepackage[english]{babel} 
\usepackage{siunitx}  %SI
\usepackage{cleveref}
\usepackage[font=normalsize]{caption}
\usepackage{url}
\usepackage{textcomp}
\graphicspath{{fig/}}
\usepackage{algorithm}
\usepackage{algpseudocode}
\usepackage{color}
\makeatletter
\newcommand{\thickhline}{%
    \noalign {\ifnum 0=`}\fi \hrule height 2pt
    \futurelet \reserved@a \@xhline
}
\newcolumntype{"}{@{\hskip\tabcolsep\vrule width 2pt\hskip\tabcolsep}}
\makeatother

\begin{document}
% Number lines
%\linenumbers
%%%%%%%%%%%%%%%%%%%%%%%%%%%%%%%%%%%%%%%%%%%%%%%%%%%%%%%%%%%%%%%%%%%%%%%%
% Cover and abstract page
%%%%%%%%%%%%%%%%%%%%%%%%%%%%%%%%%%%%%%%%%%%%%%%%%%%%%%%%%%%%%%%%%%%%%%%%

\begin{frontmatter}
\title{Automatic CNN-based detection of cardiac MR motion artefacts using k-space data augmentation and curriculum learning}

\author[KCL]{Ilkay Oksuz\corref{cor1}} 
\ead{ilkay.oksuz@kcl.ac.uk}
%\author[KCL]{Ilkay Oksuz}

\author[KCL,NHS]{Bram Ruijsink}
\author[KCL]{Esther Puyol-Ant\'on} 
\author[KCL]{James Clough} 
\author[KCL]{Gastao Cruz}
\author[KCL]{Aurelien Bustin}
\author[KCL]{Claudia Prieto}
\author[KCL]{Rene Botnar}
\author[Imperial]{Daniel Rueckert}
\author[KCL]{Julia A. Schnabel}
\author[KCL]{Andrew P. King}

\cortext[cor1]{Corresponding author at School of Biomedical Engineering \& Imaging Sciences, King\textquotesingle s College London, U.K}
\address[KCL]{School of Biomedical Engineering \& Imaging Sciences, King\textquotesingle s College London, U.K}
\address[NHS]{Guy\textquotesingle s and St Thomas\textquotesingle{} Hospital NHS Foundation Trust, London, UK.}
\address[Imperial]{Biomedical Image Analysis Group, Imperial College London, UK.}

\begin{abstract}

Good quality of medical images is a prerequisite for the success of subsequent image analysis pipelines. Quality assessment of medical images is therefore an essential activity and for large population studies such as the UK Biobank (UKBB), manual identification of artefacts such as those caused by unanticipated motion is tedious and time-consuming. Therefore, there is an urgent need for automatic image quality assessment techniques. In this paper, we propose a method to automatically detect the presence of motion-related artefacts in cardiac magnetic resonance (CMR) cine images. We compare two deep learning architectures to classify poor quality CMR images: 1) 3D spatio-temporal Convolutional Neural Networks (3D-CNN), 2) Long-term Recurrent Convolutional Network  (LRCN). Though in real clinical setup motion artefacts are common, high-quality imaging of UKBB, which comprises cross-sectional population data of volunteers who do not necessarily have health problems creates a highly imbalanced classification problem. Due to the high number of good quality images compared to the relatively low number of images with motion artefacts, we propose a novel data augmentation scheme based on synthetic artefact creation in k-space. We also investigate a learning approach using a predetermined curriculum based on synthetic artefact severity. We evaluate our pipeline on a subset of the UK Biobank data set consisting of 3510 CMR images. The LRCN architecture outperformed the 3D-CNN architecture and was able to detect 2D+time short axis images with motion artefacts in less than 1ms with high recall. We compare our approach to a range of state-of-the-art quality assessment methods.  The novel data augmentation and curriculum learning approaches both improved classification performance achieving overall area under the ROC curve of 0.89.

\end{abstract}
\begin{keyword}
Cardiac MR Motion Artefacts; Image Quality Assessment; Convolutional Neural Networks; LSTM
\end{keyword}
\end{frontmatter}

%%%%%%%%%%%%%%%%%%%%%%%%%%%%%%%%%%%%%%%%%%%%%%%%%%%%%%%%%%%%%%%%%%%%%%%%
% Introduction
%%%%%%%%%%%%%%%%%%%%%%%%%%%%%%%%%%%%%%%%%%%%%%%%%%%%%%%%%%%%%%%%%%%%%%%%

\section{Introduction}
\label{sec:introduction}

With developments in image acquisition schemes and machine learning algorithms, medical image analysis techniques are taking on increasingly important roles in clinical decision making. 
An important and often overlooked step in automated image analysis pipelines is the assurance of image quality - high accuracy requires good quality medical images. Cine  cardiac magnetic resonance (CMR) images are instrumental in assessment of cardiac health, and can be used to derive metrics of cardiac function such as volumes and ejection fractions, as well as to investigate local myocardial wall motion abnormalities.  The CMR is often often acquired for patients, who already have existent cardiac diseases, more likely to have arrythmias, have difficulties with breath-holding  or remaining still during acquisition. Therefore, the images can contain a range of image artefacts \citep{Ferreira2013}, and assessing the quality of images acquired by MR scanners is a challenging problem. Misleading conclusions can be drawn when the original data are of poor quality. Traditionally, images are visually inspected by one or more experts, and those showing an insufficient level of quality are excluded from further analysis. However, visual assessment is time consuming and prone to variability due to inter-rater and intra-rater variability.\\

\begin{figure}[tb]

\begin{minipage}[b]{0.31\linewidth}
  \centering
  \centerline{\includegraphics[width=4.0cm]{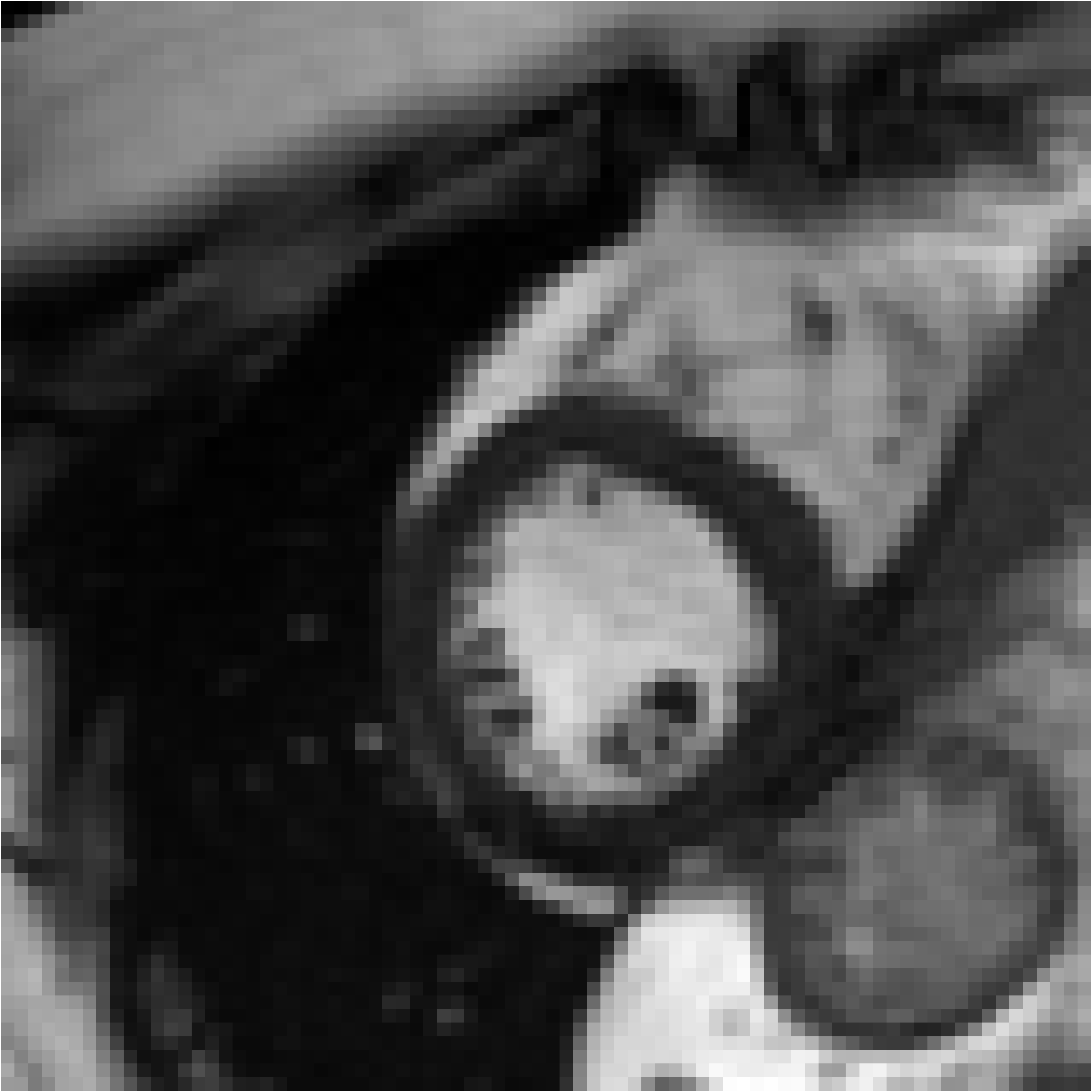}}
%  \vspace{1.5cm}
  \centerline{(a) Good quality image}\medskip
  \label{fig:Motivationa}
\end{minipage}
\hfill
\begin{minipage}[b]{0.31\linewidth}
  \centering
  \centerline{\includegraphics[width=4.0cm]{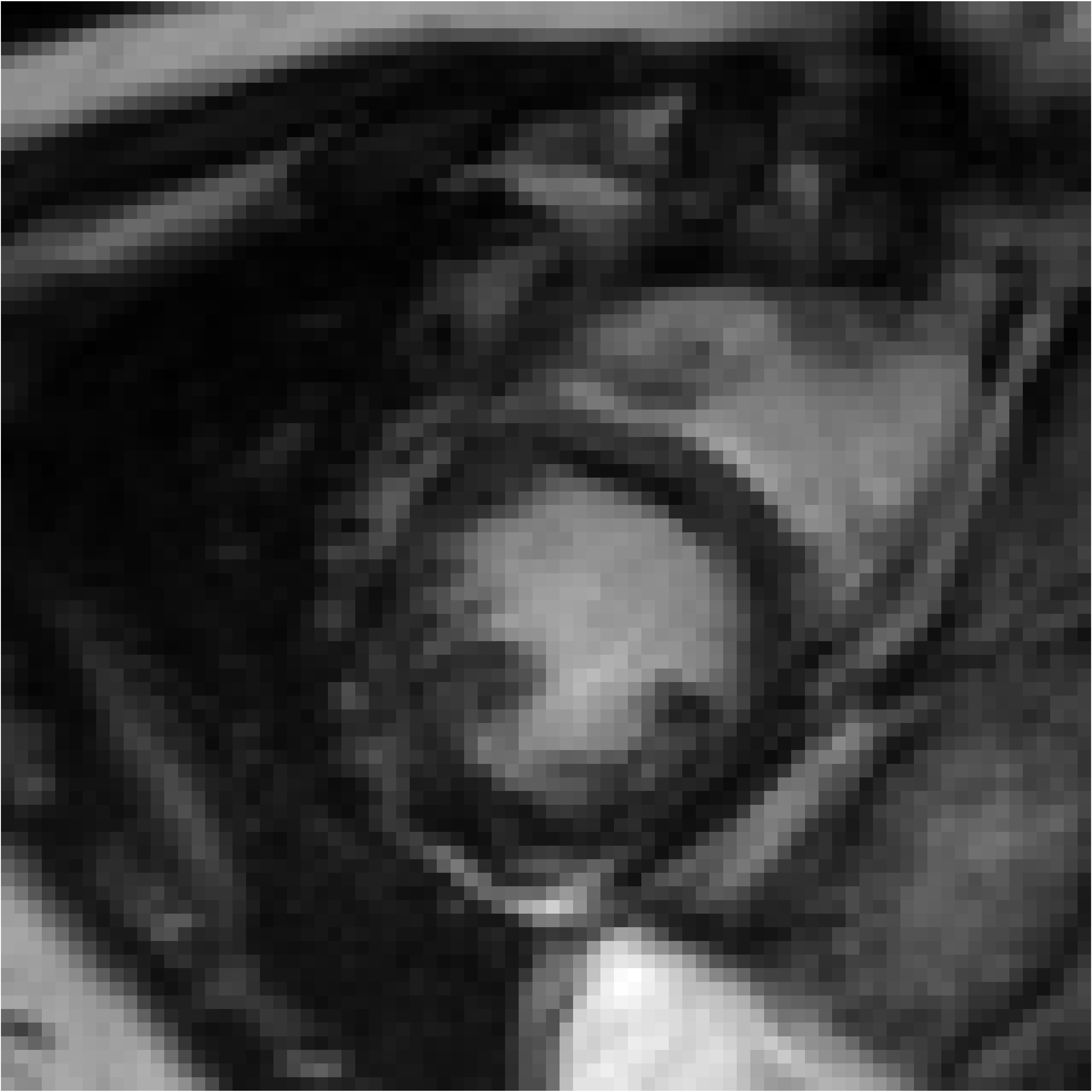}}
%  \vspace{1.5cm}
  \centerline{(b) Motion artefact image}\medskip
  \label{fig:Motivationb}
\end{minipage}
\hfill
\begin{minipage}[b]{0.31\linewidth}
  \centering
  \centerline{\includegraphics[width=4.0cm]{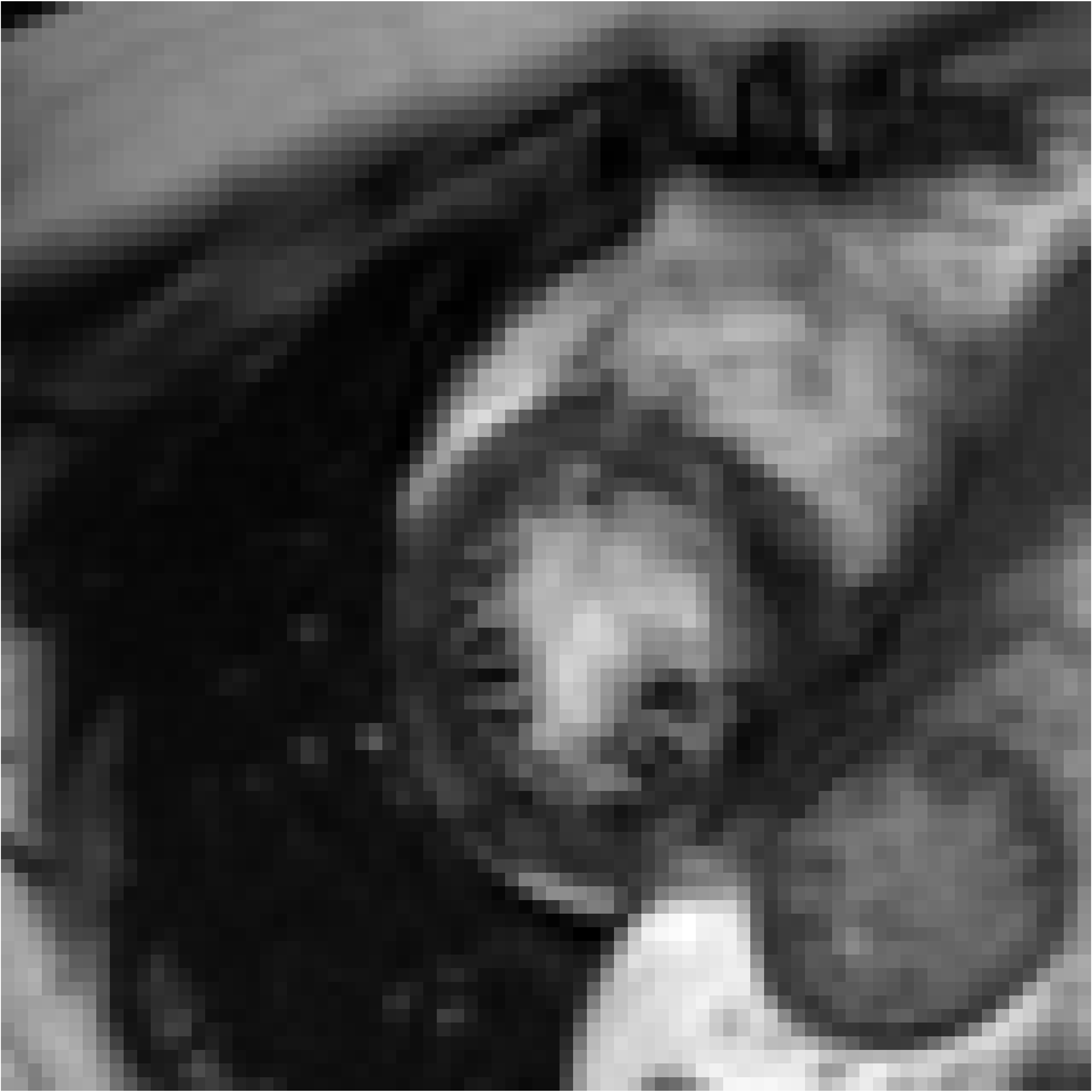}}
%  \vspace{1.5cm}
  \centerline{ (c) Synthetic image} \medskip
  \label{fig:Motivationc}
\end{minipage}
\hfill

\caption{Examples of a good quality cine CMR image (a), an image with blurring motion artefacts (b), and a k-space corrupted image (c). The k-space corruption process is able to simulate realistic motion-related artefacts. (Please see videos in supplementary material.)}
\label{fig:Mot}
\end{figure}

% UK Biobank
The UK Biobank is a large-scale study with all data accessible to researchers worldwide, and will eventually consist of CMR images from 100,000 subjects \citep{Petersen2015}. To maximise the research value of this and other similar data sets, automatic quality assessment tools are essential. One specific challenge in CMR is motion-related artefacts such as mistriggering, arrhythmia and breathing artefacts. These can result in temporal and/or spatial blurring of the images, which makes subsequent processing difficult \citep{Ferreira2013}. These type of artefacts are more common in real clinical acquisitions, and there would be great value for motion artefact detection mechanisms being deployed in the MR scanner.
For example, these types of artefact can lead to erroneous quantification of myocardial wall motion, which is an important indicator in cardiac functional assessment. Examples of a good quality image and an image with blurring motion artefacts are shown in Figure \ref{fig:Mot}a-b for a short-axis view cine CMR scan.\\ 

% Overview of paper

In this paper, we propose a deep learning based approach for fully automated motion artefact detection in cine CMR short axis images. A novel data augmentation strategy based on synthetic artefact creation in k-space and a curriculum learning scheme based on the synthetic artefacts with different levels of severity (Figure \ref{fig:MotivationCur}) is also proposed. An analysis of multiple deep learning architectures and learning mechanisms is also presented. This paper builds upon our previously presented work \citep{Oksuz2018}, in which we proposed the use of synthetically generated mistriggering artefacts in training a Convolutional Neural Network (CNN). Here, we extend this idea to include both breathing and mistriggering artefacts and also use different levels of corruption in order to produce a curriculum of realistic artefact images of varying severity (Figure \ref{fig:Mot}c) to improve training.\\

The remainder of this paper is organised as follows. In Section \ref{sec:Related}, we first present an overview of the relevant literature in image quality assessment and the data imbalance problem, which our novel extensions are based on. 
Then, we review the literature on curriculum learning, and present our novel contributions in this context. 
In Section \ref{sec:materials}, we provide details of the clinical data sets used. 
In Section \ref{sec:methods} we describe the deep learning models that we have utilised for classification, including descriptions of the novel data augmentation and curriculum learning approaches. 
Results are presented in Section \ref{sec:experiments_results}, while Section \ref{sec:discussion_conclusion} discusses the findings of this paper in the context of the literature and proposes potential future work directions.

%%%%%%%%%%%%%%%%%%%%%%%%%%%%%%%%%%%%%%%%%%%%%%%%%%%%%%%%%%%%%%%%%%%%%%%%
%Related Works
%%%%%%%%%%%%%%%%%%%%%%%%%%%%%%%%%%%%%%%%%%%%%%%%%%%%%%%%%%%%%%%%%%%%%%%%

\section{Related Works}
\label{sec:Related}

In this section, we provide an overview of the relevant literature on image quality assessment, data imbalance and curriculum learning with a focus on applications in medical image analysis.

\subsection{Image quality assessment}
\label{sec:imq}

%It is highly desirable to be able to automatically and accurately predict visual signal quality as  large amounts of data are produced. Such predictive capability can be used to monitor image and video traffic, and to improve the perceptual quality of visual signals via ‘‘quality-aware’’ processing, computing, and networking \citep{Bovik2013}.
An automatic image quality assessment (IQA) algorithm, given an input image, tries to predict its perceptual quality. The perceptual quality of an image is usually defined as the mean of the individual ratings of perceived quality assigned by human observers.
Early works on IQA focused on using Natural Scene Statistics (NSS) to predict the naturalness of the images. For example, \cite{Mittal2013} proposed the Naturalness Image Quality Evaluator (NIQE) model, which constructed a collection of statistical features based on a space domain NSS model.  
\cite{Moorthy2011} proposed a two-stage framework for estimating quality based on NSS models, involving identification- and distortion-specific quality assessment. More recently, Convolutional Neural Networks (CNNs) have been utilised for image quality assessment \citep{Kang2014} and \cite{Talebi2018} proposed a novel loss function definition and focused on the distribution of the ground truth quality scores.\\

%\subsection{Diagnostic image quality of medical images}
%\label{sec:mimq}

IQA is an essential step for analysing large medical image data sets (see \cite{Chow2016} for a comprehensive review). Early efforts in medical imaging focused on quantifying the image quality of brain MR images. \cite{Woodard2006} defined a set of 239 no-reference image-quality metrics (IQMs). However, the IQMs were calculated on image pairs with simplistic distortions such as Gaussian  noise  or  intensity  nonuniformity, which are unlikely to adequately capture the nature of real world MR image artefacts. \cite{Mortamet2009}  proposed two IQMs focused on detecting artefacts in the air region surrounding the head. They applied these IQMs in 749 scans from the Alzheimer’s Disease Neuroimaging Initiative (ADNI) database. However, many potential sources of uncontrolled variability exist between studies and sites, including MR protocols, scanning settings, participant instructions, inclusion criteria, etc. The thresholds they proposed on their IQMs are unlikely to generalise beyond the ADNI database.\\

Trends in the computer vision literature have heavily influenced medical image quality assessment techniques.
CNNs have been utilised for image quality assessment for compressed images in the computer vision literature with considerable success \citep{Kang2014}. This success has motivated the medical image analysis community to utilise them on multiple image quality assessment challenges such as fetal ultrasound \citep{Wu2017} and echocardiography \citep{Abdi2017}. These two techniques use 2D images and assess quality using pre-trained neural networks. A more recent study \citep{Abdi2017a} aimed to utilise temporal information using a Long Short Term Memory (LSTM) architecture to improve the accuracy of image quality assessment. \cite{Kuestner2018a} utilised a patch-based CNN architecure to detect motion artefacts in head and abdomen MR scans to achieve spatially-aware probability maps. In more recent work, \cite{Kuestner2018} proposed to utilise a variety of features and train a deep neural network for artefact detection. The authors made use of an active learning strategy to detect low quality images due to the lack of sufficient training data.\\

 In the context of CMR, the literature has mostly focused on  missing apical and basal slice detection \citep{Zhang2016}. Missing slices adversely affect the accurate calculation of the left ventricular volume and hence the derivation of cardiac metrics such as ejection fraction. Another study \citep{Zhang2017} used Generative Adverserial Networks in a semi-supervised setting to improve the performance of missing slice detection. 
 \cite{Tarroni2018} proposed to use a decision forest approach for heart coverage estimation, inter-slice motion detection and image contrast estimation in the cardiac region.  
 CMR image quality has also been linked with automatic quality control of image segmentation in \cite{Robinson2017}.
 \cite{Lorch2017} investigated synthetic motion artefacts and used histogram, box, line and texture features to train a random forest algorithm to detect different artefact levels. However, their algorithm was tested only on artificially corrupted synthetic data and aimed only at detecting breathing artefacts.
 %In this paper, we aim to detect CMR motion artefacts in short axis images. Our dataset contains severe data imbalance between poor and good quality images, which needs to be taken into account in algorithm development.

\subsection{Data imbalance}
\label{sec:dataimbalance}

Data imbalance is a significant factor that influences the stability of machine learning algorithms \citep{Chawla2010}. The fundamental issue with the imbalanced learning problem is the ability of imbalanced data to significantly compromise the performance of most standard learning algorithms.
%Most standard algorithms assume or expect balanced class distributions or equal misclassification costs. Therefore, when presented with imbalanced data sets, these algorithms fail to properly represent the distributive characteristics of the data and resultantly provide unfavorable accuracies across the classes of the data \citep{He2008}. \\
This occurs because the skewed distribution of class instances can lead the classification algorithms to be biased towards the majority class in classification tasks. Therefore, the features relevant to the minority class are not learned adequately. As a result, standard classifiers (classifiers that do not consider data imbalance) tend to misclassify the minority samples into majority samples, which results in poor classification performance \citep{Wang2016}.  How to deal with imbalanced data sets is a key issue in classification and it has been well explored over past decades. Until now, this issue has been solved mainly in two ways: sampling techniques and cost sensitive methods.

\subsubsection{Sampling techniques}

Sampling techniques aim to address the data imbalance problem by generating a balanced data set by sampling the full data set
%Specifically, it tries to provide a balanced distribution by over-sampling the under-represented class
\citep{Estabrooks2004}. Random over-sampling is one of the simplest sampling methods. It randomly duplicates a certain number of samples from the minority class and then augments them into the original data set \citep{Han2005}. Conversely under-sampling randomly removes a certain number of instances from the majority class to achieve a balanced data set. In general, random over-sampling may lead to overfitting while random under-sampling may result in insufficient training data.

\subsubsection{Cost sensitive learning}

In addition to sampling techniques, another way to deal with the data imbalance problem is cost sensitive learning. In contrast to sampling methods, cost sensitive learning methods solve the data imbalance problem by assigning different costs to misclassifying majority and minority samples \citep{Khan2018}. An objective function for cost sensitive learning can be constructed based on the aggregation of the overall cost on the whole training set.
%An optimal classifier can be learned by minimising the objective function.
Although cost sensitive algorithms can significantly improve classification performance, they are only applicable when the specific costs of misclassification are known. Unfortunately, in many applications a cost with appropriate weights is hard  to define \citep{Maloof2003}.

\subsubsection{Data imbalance problem for neural networks}

In the area of neural networks, many efforts have been made  to address the data imbalance problem. Nearly all of the work falls into one of the main streams mentioned above.
%In   particular,   it's   the   specific implementations  of  either  sampling  or  cost  sensitive  methods   or  their  combinations  on  neural  networks.
\cite{Zhou2006} empirically studied the effect of sampling and threshold-moving in training cost sensitive  neural  networks. Both over-sampling  and  under-sampling techniques were used to modify the distribution of the training  data  set. To avoid the potential issues with these basic approaches, a more complex sampling method was proposed. The synthetic minority over-sampling technique (SMOTE) has proven to be quite powerful and has achieved a great deal of success in various applications \citep{Han2005}. SMOTE creates artificial data based on the similarities between existing minority samples.
Our approach in this paper is related to the SMOTE approach in that we propose to generate synthetic data for the minority class using prior knowledge of the process of cine MR image acquisition.
 
\subsection{Curriculum learning}
\label{sec:curriculumintro}

% curriculum Learning computer vision
A curriculum determines a sequence of training samples, which essentially corresponds to a list of samples ranked in ascending order of learning difficulty. 
In a pioneering work \cite{Elman1993} studied the effect of a learning structure on a synthetic grammar task. His work was inspired  by  language  learning  in  children  and  demonstrated that a neural network was able  to  learn  the  grammar when training data was presented from simple to complex order and failed to do so when the order was random.\\   

The idea of learning easy things first has been an active research topic in computer vision \citep{Lee2011}. \cite{Bengio2009} demonstrated that curriculum learning resulted in better generalisation and faster learning on synthetic vision and word representation  learning  tasks.  \cite{Pentina2015}  investigated the  effect  of  curriculum  learning  in a multi-task learning setup and proposed a model to learn the order of multiple tasks. They illustrated the superiority of learning tasks sequentially instead of learning tasks jointly. \cite{Avramova2015} applied curriculum learning to a natural image classification task by training a CNN from scratch. \cite{Weinshall2018} investigated the robustness of curriculum learning in common computer vision image classification tasks and highlighted the superiority in convergence. \cite{Gui2017} proposed a curriculum learning strategy on facial expression classification.\\
%Our work differs from the previous approaches, by demonstrating the benefits  of  curriculum  learning  in  image quality assessment.  We generate gradually corrupted synthetic images to have a curricula of examples for motion artefact images. We make use of baby-step curriculum learning and introduce synthetic samples gradually to increase the robustness of the training. \\

 %In the absence of pre-defined curriculas self-paced learning have been used with considerable success. By defining loss function updates according the 

% Curriculum Learning medical images
Recently, the idea of curriculum learning has been utilised for medical imaging challenges. \cite{Jesson2017} proposed to use patches of different complexity to train a network for lung nodule detection. Their algorithm learnt how to distinguish nodules from the initial surroundings and added difficult patches gradually.
%\cite{Lotter2017} trained a network on patches and used the trained network as a feature extractor to classify images, which exhibits another setup of curriculum learning.
\cite{Maicas2018} used a teacher-student curriculum learning strategy for breast screening classification from DCE-MRI. They trained their model on simpler tasks before introducing the final problem of malignancy detection. 
\cite{Berger2018} proposed to use an adaptive sampling strategy to improve the segmentation performance on difficult regions in multi-organ CT segmentation.
%None of these models relied on a pre-defined curriculum of samples, which makes the training process prone to biases in the network. In our setup, we generate a set of gradual motion-corrupted images to train our model. \\

  \subsection{Contributions}
\label{sec:contributions}

There are three major contributions of this work:
\begin{itemize}
    \item To the authors' knowledge, this is the first paper that provides a thorough analysis of machine learning methods for automatic cine CMR motion artefact detection on a large scale in-vivo database;
    \item A synthetic data augmentation strategy is proposed using k-space corruption to simulate motion artefact data (see Figure \ref{fig:Mot}c) of varying levels of severity;
    \item A curriculum learning strategy is employed using the synthetic data to efficiently train deep learning models with training samples of increasing difficulty.
\end{itemize}
This paper builds upon our previous work \citep{Oksuz2018}, in which we proposed the use of synthetically generated mistriggering artefacts in training a CNN. Here, we extend this idea to include both breathing and mistriggering artefacts and also use different levels of corruption to enable the curriculum learning strategy to be introduced.

%%%%%%%%%%%%%%%%%%%%%%%%%%%%%%%%%%%%%%%%%%%%%%%%%%%%%%%%%%%%%%%%%%%%%%%%
% Materials
%%%%%%%%%%%%%%%%%%%%%%%%%%%%%%%%%%%%%%%%%%%%%%%%%%%%%%%%%%%%%%%%%%%%%%%%
\section{Materials}
\label{sec:materials}

%In this section, we provide an overview of the dataset we are using and describe the k-space  corruption strategy that we employ for synthetic motion artefact data generation.

%\subsection{UK Biobank data set}

We evaluate our approach using a subset of the UK Biobank data set.
The subset  consists  of  short-axis cine CMR images of 3510 subjects. This subset was chosen to be free of other types of image quality issues such as missing axial slices and was visually verified by an expert cardiologist. The short-axis images have an in-plane image resolution of $1.8 \times 1.8 $mm$^{2}$ with a slice thickness of 8.0 mm and a slice gap of 2 mm. A short-axis image stack typically consists of approximately  10 image slices and covers the full heart. Each cardiac cycle consists of 50 time frames and the full sequence of 50 balanced  steady-state free precession (bSSFP) magnitude images were used for analysis. Details of the image acquisition protocol can be found in \cite{Petersen2015}. \\

The data for the 3510 subjects consist of 3360 good quality acquisitions and 150 acquisitions with motion artefacts.
The artefact acquisitions featured 57 mistriggering artefacts, 46 breathing artefacts, 42 arrythmia artefacts and 5 mixed artefacts. Binary image quality labels were generated by visual inspection and validated by an expert cardiologist.\\

%%%%%%%%%%%%%%%%%%%%%%%%%%%%%%%%%%%%%%%%%%%%%%%%%%%%%%%%%%%%%%%%%%%%%%%%
% Methods
%%%%%%%%%%%%%%%%%%%%%%%%%%%%%%%%%%%%%%%%%%%%%%%%%%%%%%%%%%%%%%%%%%%%%%%%
\section{Methods}
\label{sec:methods}

In this section we first describe the neural network architectures used for motion artefact detection. We describe the preprocessing steps, then we detail the two network architectures used for motion artefact detection. We detail the data augmentation strategies to balance the classes and the curriculum learning setup proposed for training the network. Finally, we explain the details of the loss function and the optimisation of the networks.

\subsection{Preprocessing}
\label{sec:pre}

\begin{figure}[tb]
  \centering
  \centerline{\includegraphics[width=\linewidth]{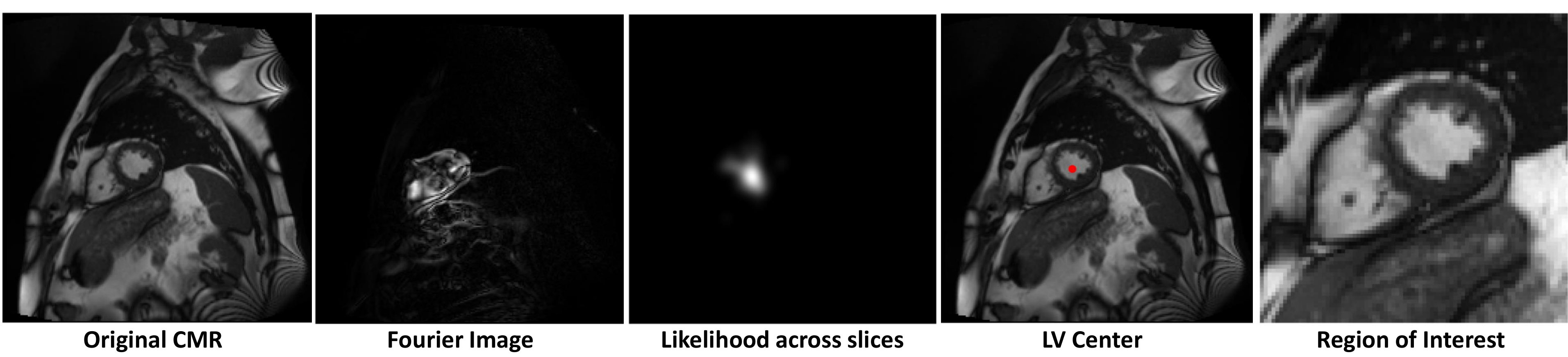}}
\caption{Region of interest extraction using Fourier transform in the temporal domain.}
\label{fig:ROI}
\end{figure}

To circumvent problems related to different image resolutions and to enable efficient memory usage we use a region-of-interest (ROI) mechanism to extract regions of consistent size (illustrated in Figure \ref{fig:ROI}). Similar to \cite{Korshunova2016}, we exploit the fact that each slice sequence captures one heart beat and use Fourier analysis to produce an image that captures the maximal activity at the corresponding heart beat frequency.  From these Fourier images, we estimate the location of the centre of the left ventricle by combining the Hough circle transform with a custom kernel-based majority voting approach across all short axis slices. First, for each Fourier image (resulting from a single slice), the  highest scoring Hough circles for a range of radii were found, and from all of those, the  top 10 highest scoring ones were retained. Finally, a ‘likelihood surface’ (centre image in Figure \ref{fig:ROI}) was obtained by combining the centres and scores of the selected circles for all slices. Each circle centre was used as the centre for a Gaussian kernel, which was scaled with the circle score, and all of these kernels were added. The maximum across this surface was selected as the centre of the ROI and $80 \times 80$ regions were extracted for further processing.  The preprocessing strategy was able to correctly identify the myocardial cavity for all cases and  was validated using the myocardial masks.\\

%The width and height of the bounding box of all circles with centers within a maximal distance (another hyperparameter) of the ROI center were used as bounds for the ROI or to create an ellipsoidal mask as shown in the figure.

\subsection{Deep learning models}
\label{sec:models}

We use deep learning methods that are capable to detect temporal dependencies in a cine sequence. In this section, we detail the two different types of video classification methods namely; 3D CNN and LRCN .\\  

\begin{figure}[tb]
  \centering
  \centerline{\includegraphics[width=\linewidth]{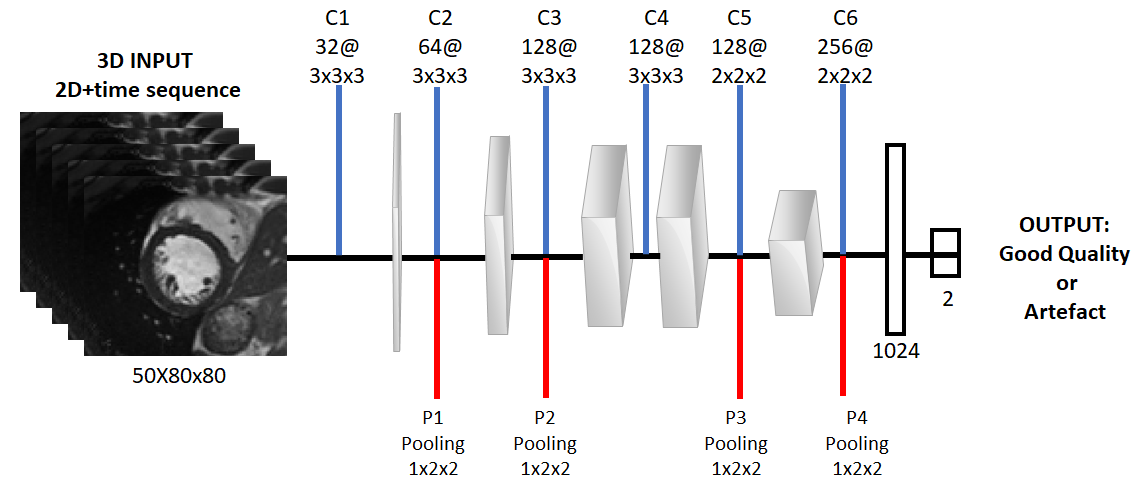}}
\caption{The 3-dimensional CNN architecture for motion artefact detection. }
\label{fig:Model}
\end{figure}

\textbf{3D CNN}: The proposed CNN consists of eight layers as visualised in Figure \ref{fig:Model}. The architecture of our network follows a similar architecture to that proposed in \cite{Tran2015}, which was originally developed for video classification using a spatio-temporal 3D CNN. In our case we use the third dimension as the time component for mid-ventricular sequences for classification. The input is an intensity normalised $80 \times 80$ cropped CMR image with 50 time frames as described in Section \ref{sec:pre}. The network has 6 convolutional layers and 4 pooling layers, 2 fully-connected layers and a softmax loss layer to predict motion artefact labels.  After each convolutional layer a Rectifier Linear Unit (ReLU) activation is used. We then apply pooling on each feature map to reduce the filter responses to a lower dimension. We  apply dropout with a probability of 0.5 at all convolutional layers and after the first fully connected layer to enforce regularisation. All of these convolutional layers are applied with appropriate padding (both spatial and temporal) and stride 1, thus there is no change in terms of size from the input to the output of these convolutional layers.

\begin{figure}[tb]

\begin{minipage}[b]{0.48\linewidth}
  \centering
  \centerline{\includegraphics[width=\linewidth]{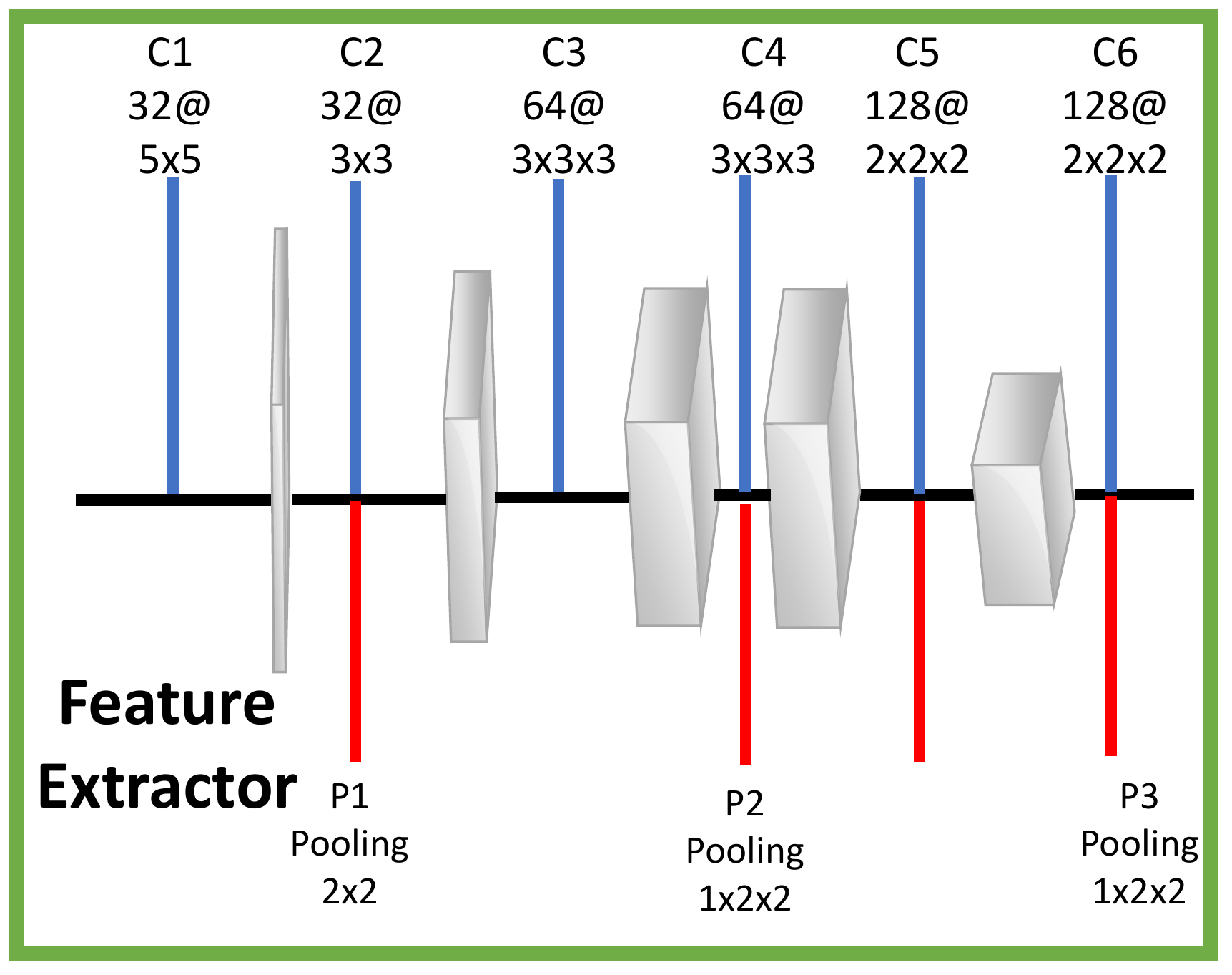}}
%  \vspace{1.5cm}
  \centerline{(a)}\medskip
\end{minipage}
\hfill
\begin{minipage}[b]{0.48\linewidth}
  \centering
  \centerline{\includegraphics[width=\linewidth]{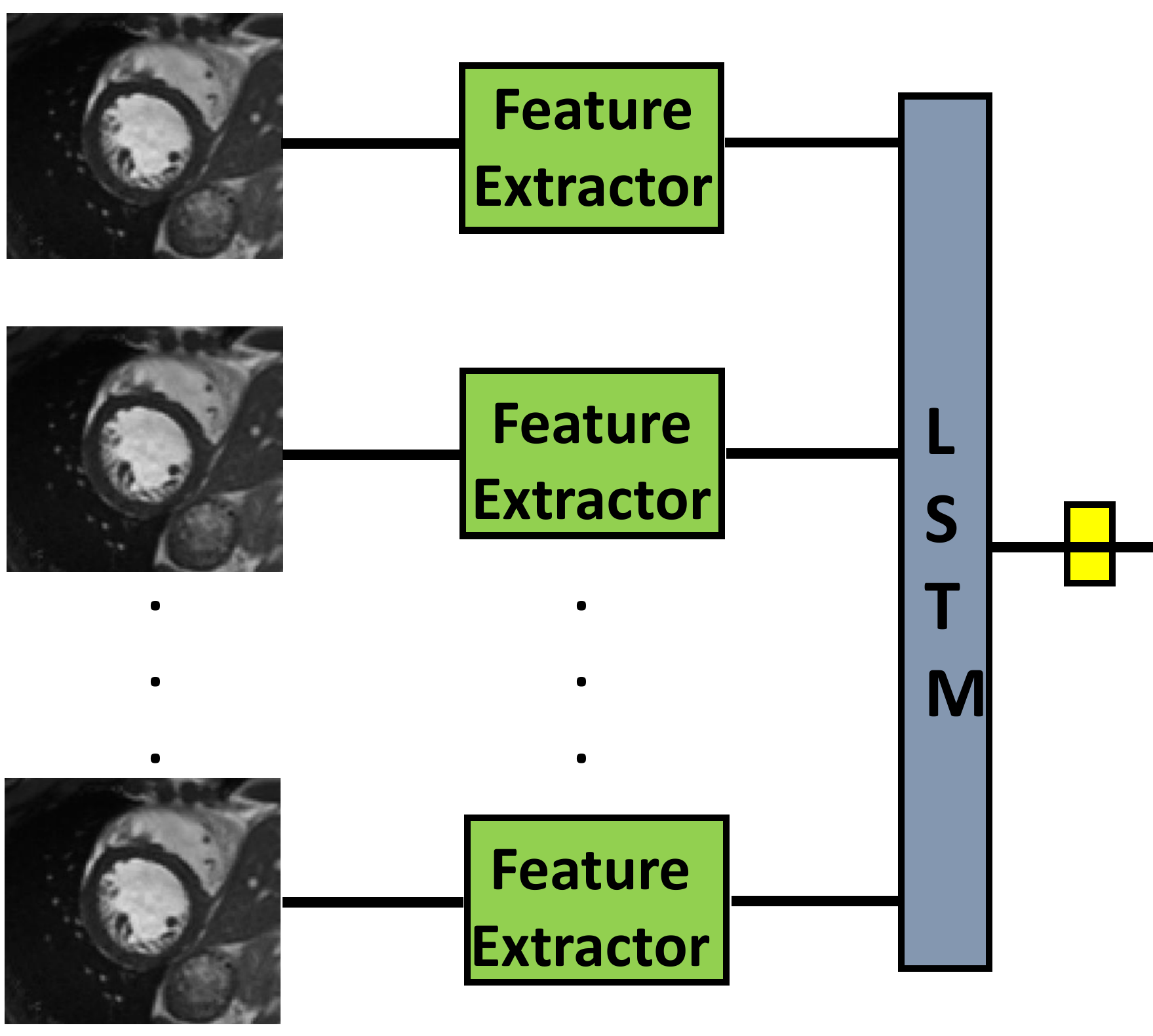}}
%  \vspace{1.5cm}
  \centerline{(b)}\medskip
\end{minipage}
\hfill

\caption{ The LRCN architecture for motion artefact detection. (a) The  feature extractor block for 2D images. (b) The network architecture. Multiple 2D inputs of different cardiac phases are passed through the feature extractor and a recurrent block (LSTM) is used for the final classification.}
\label{fig:ModelLRCN}
\end{figure}

\textbf{LRCN:} The proposed Long-term  Recurrent  Convolutional Network model follows a similar strategy to that proposed in \cite{Donahue2017}, which combines a deep hierarchical visual feature extractor (such as a CNN) with a model that can learn  to  recognise  and  synthesise  temporal  dynamics  for tasks involving sequential data. LRCN works by passing each  visual  input $x_{t}$ (an  image  in  isolation,  or  a  frame  from  a  video)  through a feature transformation $\phi$ (usually a CNN), to produce a fixed-length vector representation. In our algorithmic setup, we use a feature extractor network to produce the feature representation and pass it to a LSTM unit to make the final prediction. Figure \ref{fig:ModelLRCN}  illustrates the architecture of our network. Our feature extractor block consists of  6 convolutional layers and 3 pooling layers and vectorises the final output to be used in a recurrent fashion.

\subsection{Balancing the classes}
\label{sec:balance}

%\subsection{Synthetic data generation}
%\label{sec:Syntheticdata}
In order to address the heavy class imbalance in our data set we propose to generate synthetic artefacts using knowledge of the cine MR acquisition process. Cine CMR images are acquired using ECG triggering and typically the full k-space of one image is filled over multiple beats during a breath hold. During the acquisition mistakes with ECG-triggering can cause k-space lines to be filled with data from an incorrect cardiac phase. Similarly, breathing motion of the patient can cause k-space lines to be filled with data from a different anatomical location. We aim to simulate these mistriggering and breathing artefacts at varying levels of severity to be able to utilise a curriculum learning strategy during training.

\subsubsection{Mistriggering artefacts}

\begin{figure}[tb]
  \centering
  \centerline{\includegraphics[width=\linewidth]{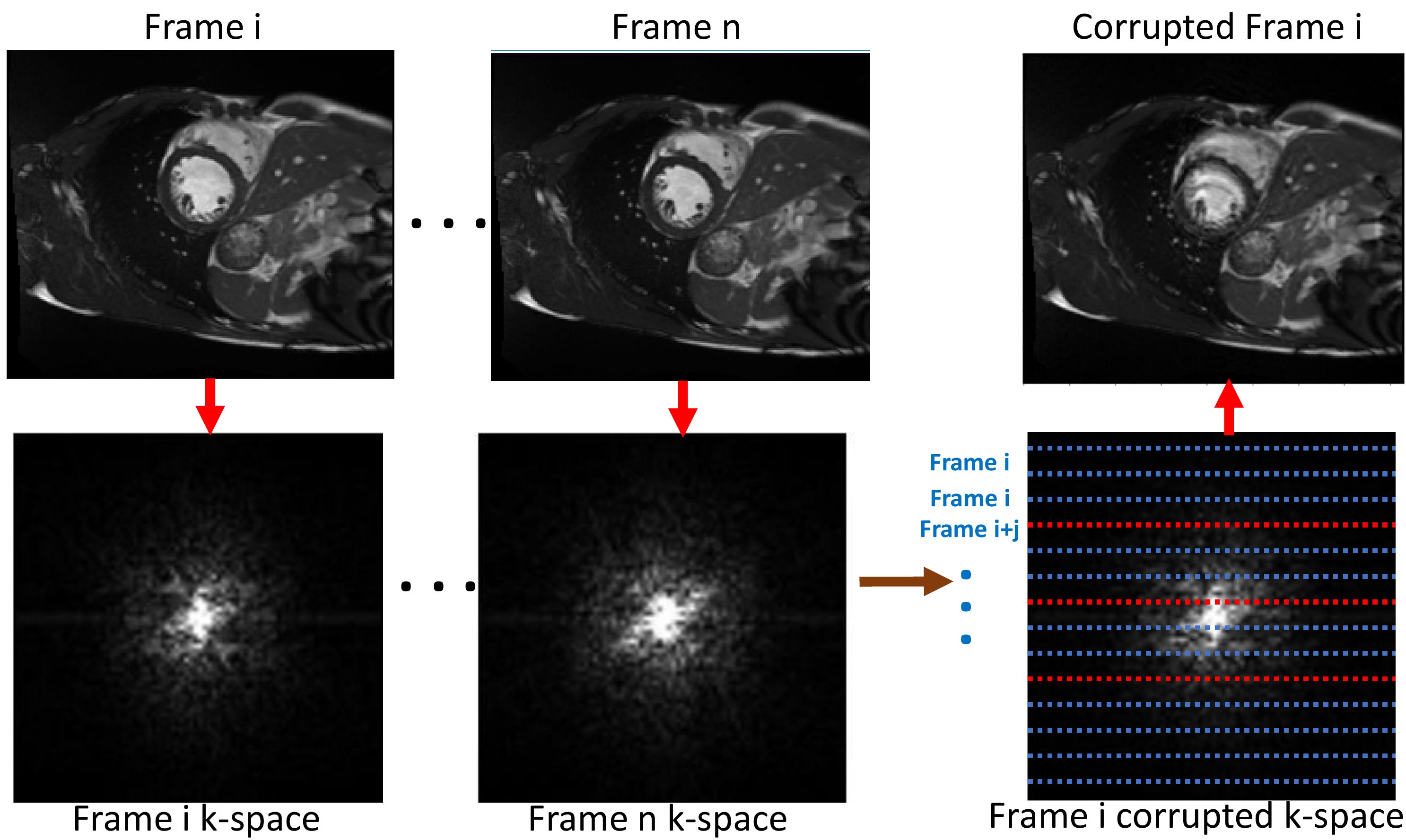}}
\caption{K-space corruption for mistriggering artefact generation in k-space. The Fourier transform of each image frame is applied to generate the k-space representation of each image. We replace k-space lines with lines from different temporal frames to generate corruptions. }
\label{fig:Kspacemistrigg}
\end{figure}

%We generate k-space corrupted data in order to increase the amount of motion artefact data for our under-represented low quality image class.
The UK Biobank data set that we use was acquired using Cartesian sampling and we follow a Cartesian k-space corruption strategy to generate synthetic but realistic motion artefacts. We first transform each 2D short axis sequence to the Fourier domain and change 1 in $z$ Cartesian sampling lines to the corresponding lines from other cardiac phases in order to mimic cardiac motion artefacts. By using different values for $z$, we are able to generate cardiac motion artefacts with different severity. In Figure \ref{fig:Kspacemistrigg} we show an example of the generation of a corrupted frame $i$ from the original frame $i$ using information from the k-space data of other temporal frames. We add a random frame offset $j$ when replacing the lines.\\

Using this approach, the original good quality images from the training set are used to increase the total number of  low quality images in the training set. This is a realistic approach as the motion artefacts that occur from mistriggering arise from similar misallocations of k-space lines.

\subsubsection{Breathing artefacts}

%We generate breathing artefacts by applying motion transformation to the frame to simulate the repository motion.
Following a similar idea to \cite{Lorch2017} we produce breathing artefacts by applying 2D translations to the image frames prior to generating their k-space representations. The translations follow a sinusoidal pattern. To simulate a subject that completed four breathing cycles within one acquisition with 256 phase-encoding steps, we sampled a sinusoidal curve with four cycles at 256 time points to produce the translations.
%A sampled point corresponds to one position in the breathing cycle; that is, a point at the bottom of the sinusoidal curve corresponds to the subject at full inhale, where as a point at the top of the sinusoidal curve corresponds to the position in the breathing cycle at full exhale.
Once the k-space representations of the frames were generated in this way they were combined in the normal way and reconstructed into images.\\

\begin{figure}[tb]
  \centering
  \centerline{\includegraphics[width=\linewidth]{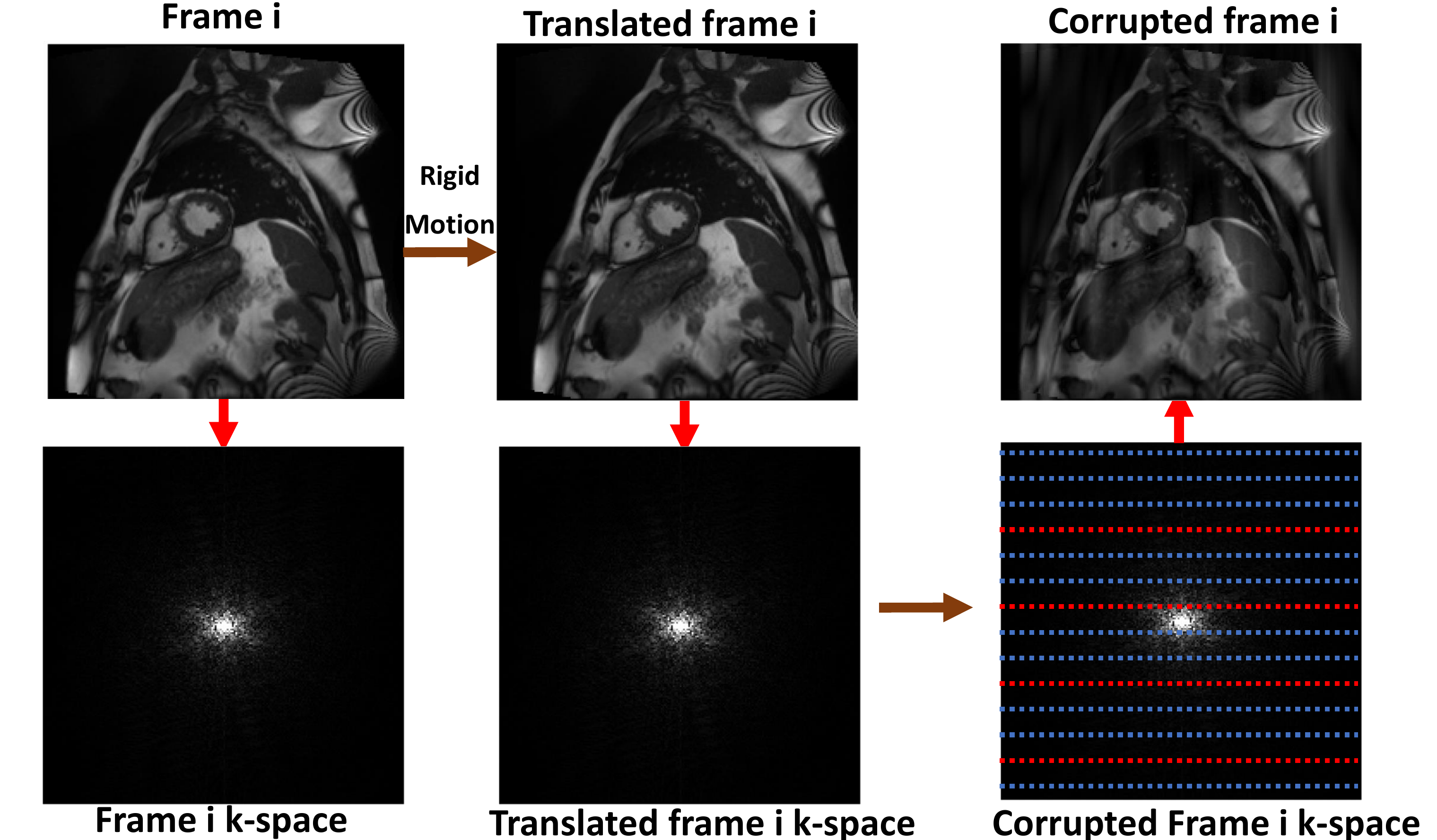}}
\caption{K-space corruption for breathing artefact generation in k-space. The Fourier transform is applied to generate the k-space of each image frame and we replace k-space lines with lines from frames with different 1D translations which follow a sinuisodal pattern to simulate repository motion. }
\label{fig:Kspacebreathing}
\end{figure}

%The strategy to generate motion artefacts synthetically uses k-space lines from the k-space of the translated images according to the sinusoidal model.
In Figure \ref{fig:Kspacebreathing} we show an example of the generation of a corrupted frame $i$ from the original frame $i$ using information from the k-space data of other translated frames \citep{Cruz2016}.

%For periodic motion, however, the subsequent lines in k-space came from different images depicting the subject at subsequent positions in the breathing cycle according to the sinusoidal model.

\subsection{Curriculum learning}
\label{sec:curriculum}

We propose to use baby-step\footnote{The term baby-step refers to keeping the previously introduced training samples in the pool of training examples rather than replacing them with new ones.} curriculum learning during training of the networks to leverage the additional data resulting from the k-space corruption strategy. We start the network training with heavily corrupted images (easy examples) and gradually introduce less corrupted images (hard examples). \\

Formally, we have a training data set of images $D={(I_{1},y_{1}), \dots, (I_{n},y_{n})} $, where $I_{i} \in R^{d}$ denotes the $i^{\text{th}}$ cardiac sequence of training samples, $y_{i}$ represents its label and $n$ is the number of training samples. The estimated label $\hat{y_{i}}$  is predicted by $f(I_{i},W)$, where $W$ represents the model parameters of the decision function $f$. Let $L(y_{i}, f(I_{i}, W))$ denote the loss function which calculates the cost between the ground truth label $y_{i}$ and the estimated label $\hat{y_{i}}=f(x_{i},W)$. The motion artefact detection is then optimised by:

$$ W^{*}= \underset{W}{\operatorname{argmin}} \sum_{i=1}^n L(y_{i}, f(I_{i}, W)) $$

\begin{figure}[tb]
  \centering
  \centerline{\includegraphics[width=\linewidth]{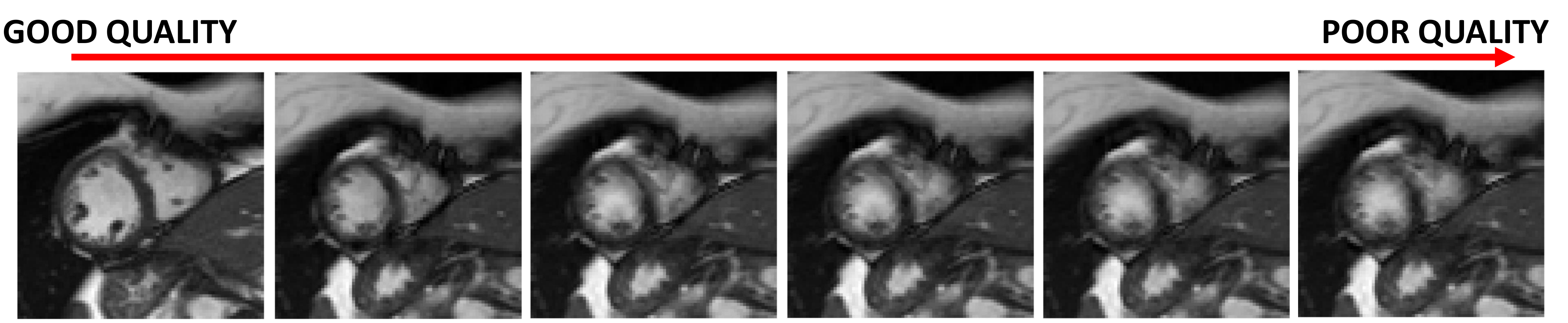}}
\caption{Gradual corruption using mistriggering type synthetic artefact generation for curriculum learning. The myocardial borders and papillary muscles become more blurred with the severity of the artefacts and it is harder to distinguish those structures under severe artefact cases.}
\label{fig:Gradual}
\end{figure}

\begin{figure}[tb]

\begin{minipage}[b]{0.38\linewidth}
  \centering
  \centerline{\includegraphics[width=\linewidth]{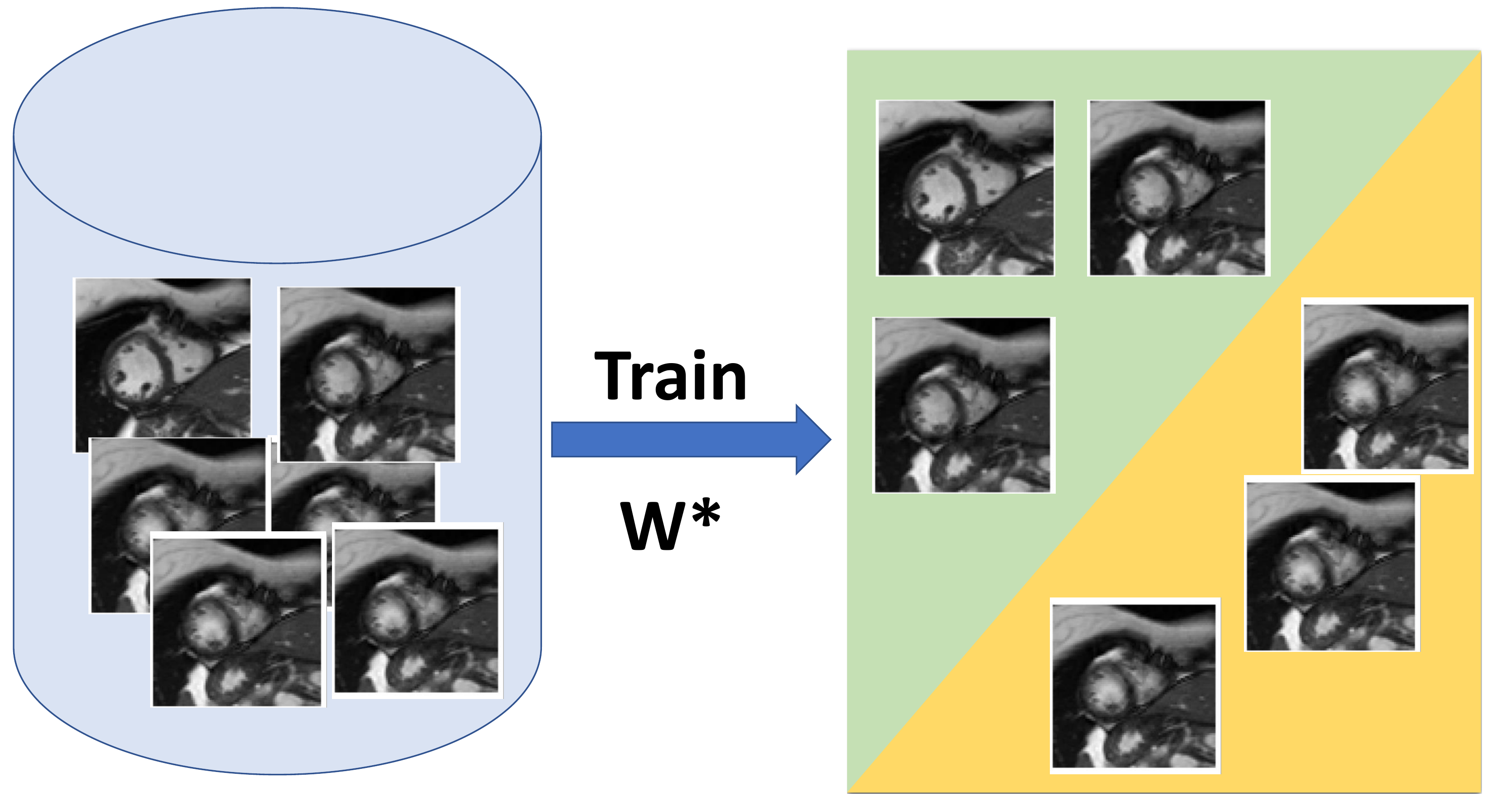}}
%  \vspace{1.5cm}
  \centerline{(a)}\medskip
\end{minipage}
\hfill
\begin{minipage}[b]{0.57\linewidth}
  \centering
  \centerline{\includegraphics[width=\linewidth]{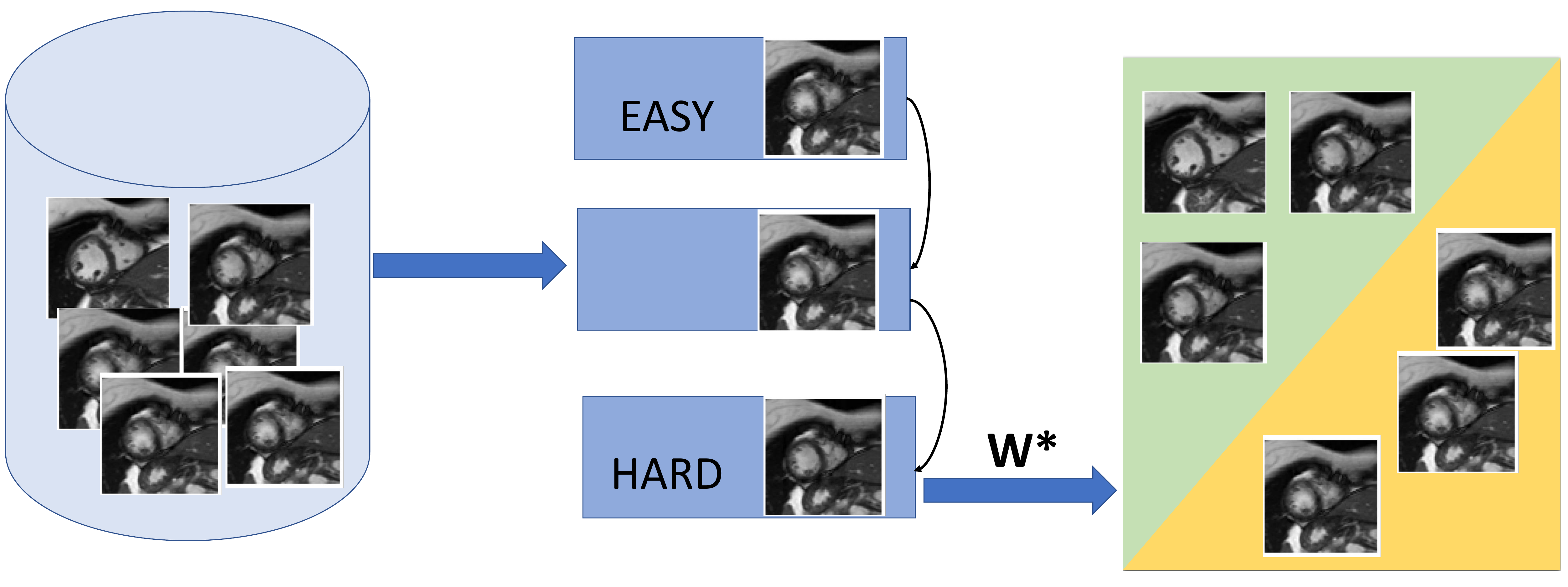}}
%  \vspace{1.5cm}
  \centerline{(b)}\medskip
\end{minipage}
\hfill

\caption{Curriculum learning using motion artefacts generated with various levels of severity. (a) The  traditional  way  to  train  a  model  fails  to consider  the  complexity  of  image quality detection  where  introducing  noisy  or  difficult  samples  early  in  training  may impair model performance. (b) The training data is divided into  different  difficulty  levels  based  on  a  predetermined curriculum. The training procedure progresses from easy to hard image  samples, which guides the model to achieve better performance.  (The  illustration  of  complexity  is  shown  in Figure \ref{fig:Gradual}).}
\label{fig:MotivationCur}
\end{figure}

Here, $W^{*}$ denotes the optimal model parameters. We utilise the different levels of corruption achieved by the k-space corruption strategy (as visualised in Figure \ref{fig:Gradual}) to sort the training samples according to their difficulty for classification.  This leads to the proposed algorithm illustrated in Figure \ref{fig:MotivationCur}. We first group image sequences into subsets based on the corruption level of the poor quality image (i.e., from high level of corruption to low level of corruption). We then train the model via iterative learning using increasingly corrupted images as described in Algorithm \ref{alg:algo1}.

\begin{algorithm}[t]
\caption{Proposed curriculum learning strategy for motion artefact detection}\label{alg:algo1}

\begin{flushleft}
        \textbf{INPUT:}  Data set of synthetically generated image sequences $D=\{ D^{i}\}^{b}_{i=1} $ ordered by a pre-defined curriculum \\
        \textbf{OUTPUT:} Optimized model parameters $W^{*}$ 
\end{flushleft}

\begin{algorithmic}[1]

\State $D^{\text{train}}= \text{Original Data set of Image Sequences}$

\For{i=\{1,\dots,b\}}

\State $D^{\text{train}}=D^{\text{train}} \bigcup D^{i}$

\For{epoch=\{1,\dots,k\}}

\State train $(W, D^{train})$
\EndFor

\State select best  $W^{*}$ 
%\State update learning rate
\EndFor

\end{algorithmic}
\end{algorithm}

\subsection{Loss functions and optimisation}
\label{sec:implementationdetails}

The training of a CNN can be viewed as a combination of two components: a loss function or training objective, and an optimisation algorithm that minimises this function. In this study, we use the stochastic gradient descent (SGD) optimiser  to minimise the binary cross entropy. The cross entropy represents the dissimilarity of the approximated output distribution from the true distribution of labels after a softmax layer and is defined as:

$$L= \dfrac{-1}{n} \sum_{i=1}^n  [y_{i}~log(\hat{y_{i}}) + (1-y_i)log(1-\hat{y_i}) ] $$

The training converges when the network does not significantly improve its performance on the validation set for a predefined number of epochs (100). An improvement is considered sufficient if the relative increase in performance is at least 0.5\%.  \\

During training, a batch-size of 50 2D+time sequences was used due to memory constraints. The  momentum of the optimiser was set to 0.90 and the learning rate was 0.0001. The parameters of the convolutional and fully-connected layers were initialised from a zero mean, unit standard deviation gaussian distribution. In each trial, training was continued until the network converged. Convergence was defined as a state in which no substantial progress was observed in the training loss. Parameters were optimised using a grid-search among all parameters. We used the Keras Framework with Tensorflow backend for implementation and training the network with the curriculum learning setup took around 12 hours on a NVIDIA Quadro 6000P GPU. Classification of a single 2D+time image  sequence took less than 1s.

%%%%%%%%%%%%%%%%%%%%%%%%%%%%%%%%%%%%%%%%%%%%%%%%%%%%%%%%%%%%%%%%%%%%%%%%
% Experiments and Results
%%%%%%%%%%%%%%%%%%%%%%%%%%%%%%%%%%%%%%%%%%%%%%%%%%%%%%%%%%%%%%%%%%%%%%%%
\section{Experiments and results}
\label{sec:experiments_results}

Three sets of experiments were performed. The first set of experiments (Section \ref{sec:SyntheticResults}) aimed to compare the performance of the different algorithmic approaches for automatic motion artefact detection, while the second set of experiments (Section \ref{sec:Quantitative}) aimed at comparing different design choices for balancing the classes. Finally, the experiments in Section \ref{sec:CurResults} validate the proposed curriculum learning training strategies.
All experiments were carried out using the Python programming language, using standard Python libraries, Tensorflow, Keras and the scikit-learn Python toolkit \citep{Pedregosa2011}.
%Section \ref{sec:error_measures} details the evaluation measures and methods of comparison used for the validation of the proposed algorithms, Section \ref{sec:SyntheticResults} describes the synthetic set of experiments and Section \ref{sec:Quantitative} presents the quantitative results on in-vivo dataset. Finally, we report the influence of curriculum learning mechanism in Section \ref{sec:CurResults}. \\
Before describing the experiments in detail, we first describe the evaluation measures used.\\

\subsection{Evaluation metrics and methods of comparison}
\label{sec:error_measures}
A 10-fold repeated stratified cross-validation was used to validate the performance of each algorithm. In each fold, the classification accuracy (i.e. the proportion of subjects correctly classified), as well as the recall (the proportion of artefact images correctly classified) and the precision (the proportion of correctly classified good quality images) were computed. Finally, we computed the average balanced accuracies and the area under the ROC curve (AUC).
%We do not report accuracy scores as they are not sufficiently discriminative in performance evaluation of heavily imbalanced datasets.
%We report a variety of measures to highlight the classification performance under severe data imbalance.
The accuracy, precision, recall, balanced accuracy and AUC metrics are defined as:
 
$$\text{Accuracy}= \dfrac{TP+TN}{TP+FP+FN+TN}$$
$$\text{Precision}= \dfrac{TP}{TP+FP} \quad  \quad  \text{Recall}= \dfrac{TP}{TP+FN}$$
$$\text{Balanced Accuracy}= \dfrac{\text{Precision+Recall}}{2}$$ 
$$\text{AUC}= \int_{-\infty }^{\infty } \text{TPR}(t) \text{FPR}(t) dt$$ \\

where $TP$ represents true positives, $FP$ is false positives, $FN$ is false negatives and $TN$ is true negatives. TPR defines the true positive rate and FPR defines the false positive rate for a given threshold $t$.\\
%In our experiments this threshold is probability of belonging to any of the two classes.\\

% Machine Learning Methods of comparison
We compared our algorithm with a range of alternative classification techniques: K-nearest neighbours, Support Vector Machines (SVMs), Decision Trees, Random Forests, Adaboost  and  Naive Bayesian. The inputs to all algorithms were the cropped intensity-normalised data as described in Section \ref{sec:pre}. We optimised the parameters of each comparative algorithm using a grid search. We also tested two techniques developed for image quality assessment in the computer vision literature: the NIQE metric \citep{Mittal2013} is based on natural scene statistics and was trained using a set of good quality CMR images to establish a baseline for good image quality; and the Variance of Laplacians is a moving filter that has been used to detect the blur level of an image. For both of these techniques we used a SVM for classification of the final scores.

\subsection{Synthetic data}
\label{sec:SyntheticResults}

We first tested our algorithm using synthetically generated artefacts to evaluate its performance. We generated different levels of corruption from good quality images using the pipeline explained in Section \ref{sec:balance} and evaluated the algorithms on a balanced data set consisting of 3360 good quality and 3360 artefact images with different severity. We used a  10-fold cross validation to classify the good quality and artefact images. The results are reported in Tables \ref{table:synmis} and \ref{table:synbreath}  for breathing and mistriggering artefacts respectively.  The high performance of the deep learning architectures is evident for both types of artefact. 
LRCN and 3D-CNN show the highest performance in terms of accuracy, recall and balanced accuracy. The general high performance by all methods can be explained by the low complexity of the problem (i.e. original vs. synthetically corrupted version of the same image) and the availability of the balanced data set.

% Synthetic Mistriggering Table
\begin{table} 
\centering
\caption{Mean accuracy (A), precision (P), recall (R) and balanced accuracy (BA) results of image classification on synthetic mistriggering artefact data. A 10-fold cross validation was used and each image was labelled once over all folds.}
\begin{tabular}{lcccc}
\hline
Methods    & A & P & R  & BA\\
\hline 

K-Nearest Neighbours  & $0.742$  & $0.742$  & $0.746$    & $0.744$     \\
Linear SVM    & $ 0.748 $  & $0.743$  & $0.749$  & $0.746$   \\
Decision Tree   &   $ 0.756 $  & $0.757$  & $0.751$  & $0.754$   \\
Random Forests   & $0.787$    & $0.782$  & $0.786$   & $0.784$  \\
Adaboost   & $0.783 $  & $0.781$  & $0.778$       & $0.779$   \\
Naive Bayesian   & $0.809$  & $0.796$  & $0.804$    & $0.800$     \\
Variance of Laplacian  & $0.802$  & $0.799$ & $0.803$   & $0.802$   \\
NIQE \citep{Mittal2013}  & $0.922$  & $0.919 $& $0.925$  & $0.923$     \\
\hline
\textbf{3D CNN }    & 0.961 & 0.957 & 0.959  & 0.958     \\
\textbf{LRCN }   & \textbf{0.963} &  \textbf{0.963} & \textbf{0.965} &  \textbf{0.964}  \\

\hline
\end{tabular}
\label{table:synmis}
\end{table}

% Synthetic Breathing Table

\begin{table} 

\centering
\caption{Mean  accuracy (A), precision (P), recall (R) and balanced accuracy (BA) results of image classification on synthetic breathing artefact data. A 10-fold cross validation was used and each image was labelled once over all folds.}
\begin{tabular}{lcccc}
\hline
Methods    & A & P & R  & BA\\
\hline 

K-Nearest Neighbours  & $0.718$  & $0.724$  & $0.721$    & $0.723$     \\
Linear SVM    & $ 0.74 $  & $0.737$  & $0.744$  & $0.741$   \\
Decision Tree   &   $ 0.707 $  & $0.708$  & $0.713$  & $0.711$   \\
Random Forests   & $0.764$    & $0.776$  & $0.781$   & $0.778$  \\
Adaboost   & $0.768 $  & $0.768$  & $0.772$       & $0.770$   \\
Naive Bayesian   & $0.788 $  & $0.790$  & $0.797$    & $0.793$     \\
Variance of Laplacian  & $0.809$  & $0.820$ & $0.824$   & $0.822$   \\
NIQE \citep{Mittal2013}  & $0.897$  & $0.899 $& $0.904$  & $0.902$     \\
\hline

\textbf{3D CNN}    & $0.953$ & $0.951$ & $0.961$ &  $0.955$     \\
\textbf{LRCN}   & \textbf{0.961} & \textbf{0.962} & \textbf{0.964} &  \textbf{0.963}     \\

\hline
\end{tabular}
\label{table:synbreath}
\end{table}

\subsection{Augmentation technique analysis}
\label{sec:Quantitative}

We evaluated deep learning algorithms on the real in-vivo cases using 150 artefact and 3360 good quality  images. We tested six different training configurations of two neural network strategies to evaluate their performance in more detail: (1) training using only acquired magnitude data without any data augmentation, (2) training using data augmentation with translations only, (3) training using data augmentation with mistriggering k-space corrupted data only, (4) training using data augmentation with breathing k-space corrupted data only, (5) training using both mistriggering and breathing type synthetic artefacts, (6) cost-sensitive learning with a weighted cost function.  The cost sensitive learning used a weighted binary cross entropy loss function with the weights determined by the ratio of samples in the classes ($150:3360$). We also augmented data in this setup using translations for a fair comparison, but the data augmentation was not used to balance the classes in this scenario and the augmentation was applied in the same way for both classes. In each setup, the acquired data corrupted by motion artefacts were used together with the real motion artefact data.\\

The translational data augmentation used random shifts in both the horizontal and vertical directions in the range of [W/5, H/5], where W and H represent the width and height of the image respectively (i.e. W=H=80 pixels= 144 mm in our case). Rotations were not used due to their influence on image quality caused by the necessary interpolation. Note that none of the augmented data were used for testing. They were only used for increasing the total number of training images.\\

We used a 10-fold stratified cross validation strategy to test all algorithms, in which each image appeared once in the test set over all folds. Due to the high class imbalance, all algorithms achieved over 0.9 accuracy and so we do not report this metric in Table \ref{table:quan}. %We do not rely only on accuracy in our results due to the bias introduced by the imbalanced classes.
The interesting comparison for the methods lies in the recall numbers, which quantify the capability of the methods to identify images with artefacts. The results show that the LRCN-based technique is capable of identifying the presence of motion artefacts with high recall compared to the other techniques.

\begin{table} 

\centering
\caption{Mean balanced accuracy (BA), precision (P), recall (R) and area under the ROC curve AUC) results of image classification for motion artefacts (in-vivo data set) trained on real and synthetic data sets. A 10-fold cross validation was used and each image was labelled once over all folds. t-aug, m-aug, b-aug represent translational, mistriggering and breathing type augmentations respectively. b-m-aug  represents a random mix of  mis-tiggering and breathing artefacts to balance the data set. cs stands for cost-sensitive learning with weighted losses.}
\begin{tabular}{lcccc}
\hline
Methods   &  BA & P & R & AUC \\
\hline 

3DCNN no-aug    & $0.590 $ & $0.713$  & $0.467$    & $0.581$    \\
3DCNN t-aug     & $ 0.679 $  & $0.751$  & $0.607$  & $0.674$    \\
3DCNN m-aug     &  0.717 & 0.762 & 0.673 & 0.732  \\
3DCNN b-aug     &  0.695 & 0.703 & 0.687 & 0.699  \\  
3DCNN cs        &  0.515 & 0.503 & 0.520& 0.613 \\
3DCNN b-m-aug   &  0.721 & \textbf{0.768} & 0.673 & 0.735    \\
\hline
LRCN no-aug             & $0.629 $ & $0.724$  & $0.533$    & $0.603$    \\
LRCN t-aug              & $0.664 $ & $0.722$  & $0.607$    & $0.704$    \\
LRCN m-aug               &  $ 0.731  $ &   $ 0.743 $  &   $ 0.720 $ &   $0.826$  \\
LRCN b-aug              & $0.719 $ & $0.731$  & $0.707$    & $0.759$     \\
LRCN cs                 &  0.511 & 0.502 & 0.520 & 0.608 \\
\textbf{LRCN b-m-aug}   & \textbf{0.742}  &  0.751 &  \textbf{0.733} &  \textbf{0.828}  \\

\hline
\end{tabular}
\label{table:quan}
\end{table}

\subsection{Influence of curriculum learning}
\label{sec:CurResults}

We investigated the influence of curriculum learning on our algorithm. For these experiments we used the best performing model from Table \ref{table:quan}, namely the LRCN model with a mixture of breathing and mistriggering synthetic artefacts.  During generation of the synthetic training samples we used $b=10$ different levels of k-space corruption and used these to generate the curriculum. We introduced the easy samples (highly corrupted images) at the beginning of the training and gradually included harder samples (less corrupted images).\\

\begin{figure}[tb]
  \centering
  \centerline{\includegraphics[width=\linewidth]{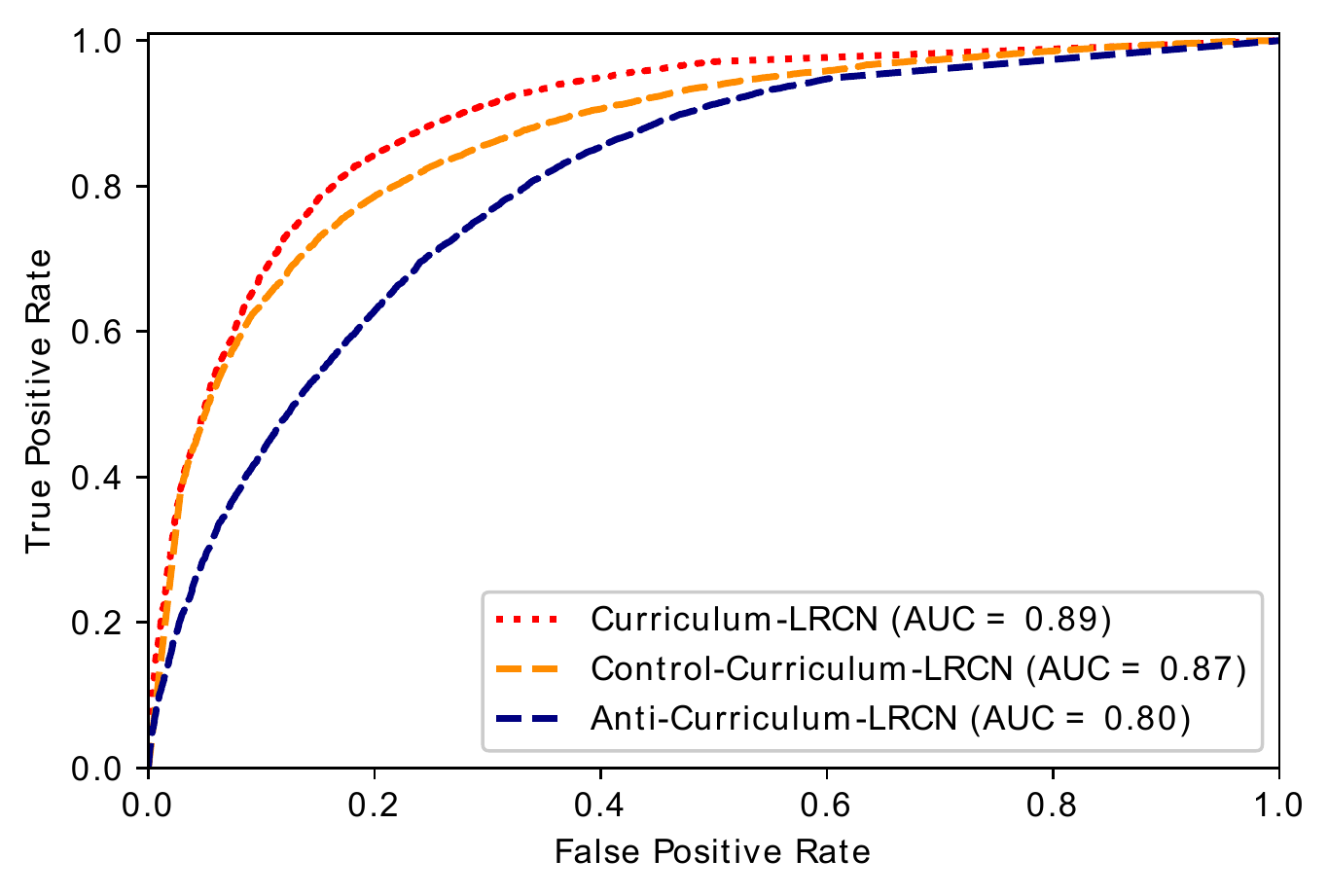}}
\caption{ROC curves for the LRCN-based motion artefact detection approach using curriculum learning. Gradually introducing harder samples during training improves the performance of the algorithm compared to the random (control-curriculum) and harder-to-easier configurations (anti-curriculum).}
\label{fig:Curresult}
\end{figure}

%This setup  is defined as curriculum learning in our experiments.
In order to evaluate the success of this approach, we compared to two alternatives. First, we repeated the curriculum generation process in the opposite way and first used hard samples and gradually introduced easier samples (anti-curriculum). Second, we used a curriculum consisting of a random set of samples with no sorting at each run (control-curriculum). Figure \ref{fig:Curresult} shows ROC curves and reports AUC values for these three approaches. We performed a Delong's statistical significance test \citep{Delong1988} to evaluate the differences between the methods. The curriculum learning strategy significantly outperformed random sampling and anti-curriculum learning (p-value $ < 0.05$).\\

\begin{figure}

\begin{minipage}[b]{\linewidth}
  \centering
  \centerline{\includegraphics[width=\linewidth]{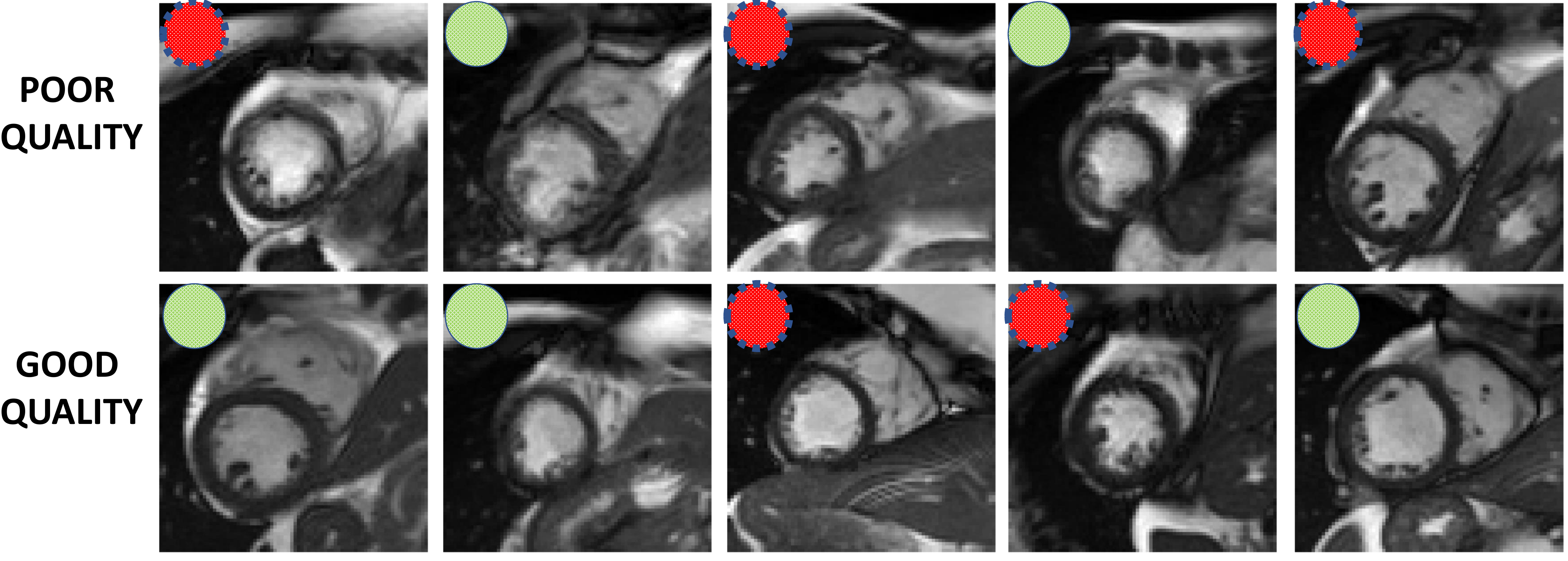}}
%  \vspace{1.5cm}
  \centerline{(a) LRCN-Control Curriculum}\medskip
  \label{fig:Cursamplesa}
\end{minipage}
\hfill
\begin{minipage}[b]{\linewidth}
  \centering
  \centerline{\includegraphics[width=\linewidth]{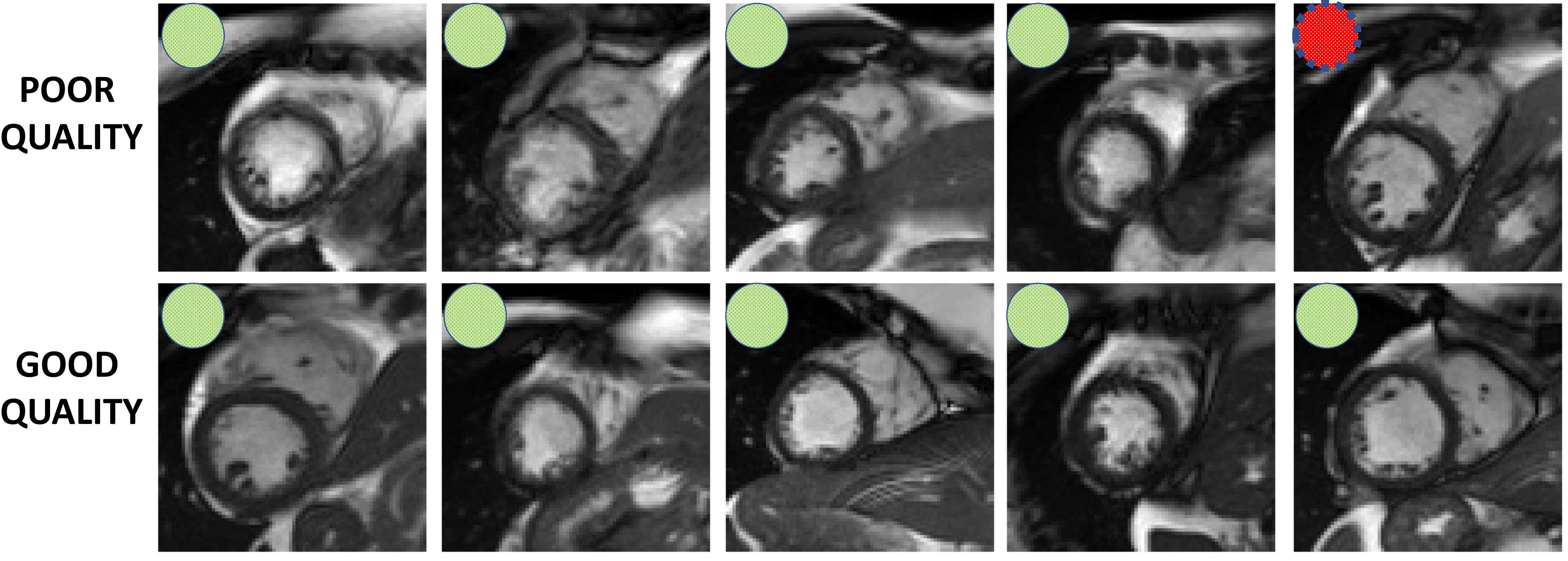}}
%  \vspace{1.5cm}
  \centerline{(b) LRCN Curriculum}\medskip
  \label{fig:Cursamplesb}
\end{minipage}
\hfill

\caption{Curriculum learning improves the classification of motion artefacts on borderline cases. (a) shows the results of control-curriculum with coloured circles for good and poor quality images.  (b) illustrates the results of the curriculum learning  strategy for the same samples. Most of the borderline cases are correctly identified with the curriculum learning strategy. The green circles indicate the correct classifications and red circles indicate the wrong classifications by the methods.}
\label{fig:Cursamples}
\end{figure}

We show the improvements in classification using curriculum learning using samples from the data set in Figure \ref{fig:Cursamples}. Some difficult classification cases were selected to showcase the performance of both methods. The figure shows the borderline cases from both classes, where there is only a slight difference between the good and poor quality images. The use of curriculum learning enables detection of borderline cases of motion artefacts (poor quality images) with great success compared to control-curriculum.

%%%%%%%%%%%%%%%%%%%%%%%%%%%%%%%%%%%%%%%%%%%%%%%%%%%%%%%%%%%%%%%%%%%%%%%%
% Discussion
%%%%%%%%%%%%%%%%%%%%%%%%%%%%%%%%%%%%%%%%%%%%%%%%%%%%%%%%%%%%%%%%%%%%%%%%
%\newpage
\section{Discussion and conclusion}
\label{sec:discussion_conclusion}

% Work summary
We have presented an extensive study on automatic cardiac motion artefact detection using spatio-temporal deep learning techniques. The motion artefact detection problem exhibits a severe data imbalance between the classes in UKBB dataset. Our fundamental contribution in this paper is to address this data imbalance by using a k-space based corruption strategy to increase the robustness of the classification. With a variety of synthetic data generation techniques we propose to augment data for training the classifier using knowledge of the cine MR acquisition process. We have also investigated the robustness of two deep learning architectures developed for video classification for classifying motion-related artefacts.
%We synthetically generate breathing and mistriggering artefacts with different levels of severity.
Benefiting from the controlled environment of synthetic data generation we utilised curriculum learning for training and showcased the efficiency of the technique in comparison with other data sampling strategies. \\

% Results comment
One key observation of our work is the superiority of deep learning methods to classify motion artefacts compared to other state-of-the-art machine learning algorithms. Moreover, we tested data augmentation strategies extensively and illustrated the superior performance of k-space corruption to generate synthetic data for augmentation. It is interesting to observe that using different corruption strategies improves the performance of the classification techniques. Finally, employing a curriculum learning strategy for training the image classification networks ensured better performance compared to anti-curriculum and random sampling strategies.\\

% Limitations

In future, we would like to investigate novel loss functions for the detection of image quality. Moreover, investigation of basal and apical slice quality, which exhibits a slightly different anatomy and challenge, is an important future direction. In this work, we deliberately used existing network architectures and loss functions to enable us to focus our evaluation on the influence of our novel data augmentation and curriculum learning strategies. In future work we would like to investigate novel architectures tailored to the problem at hand. \\

%  Clinical impact and Future work

The UK Biobank is a controlled study and the number of images with motion artefacts is limited. In real clinical acquisitions the likelihood of motion artefact occurrence is higher (although the classes would still be imbalanced), and there would be great value for motion artefact detection mechanisms being deployed `on-the-fly' in the MR scanner. The indication for CMR is often prognostic stratification of already existent cardiac diseases and  patients are more likely to have arrythmias, have difficulties with breath-holding  or remaining still during acquisition. With the successful translation of such tools in clinical setups high diagnostic image quality could be ensured on the spot. Indeed, these mechanisms would not necessarily need to be CMR specific and could even be applied to different medical image modalities and different organs.\\
%In the future, we would like to validate our algorithmic setup on different datasets and image modalities.  \\

% Conclusion 
In conclusion, we believe that the work that we have presented represents an important contribution to the understanding of CMR image quality assessment. Our novel ideas of leveraging k-space corruption for data augmentation and training the classifier with a curriculum learning strategy have been shown to improve motion artefact detection accuracy.  In the current environment of the increasing use of imaging in clinical practice, as well as the emergence of large population data cohorts which include imaging, our proposed automated quality control methods can ensure the accuracy of subsequent analysis pipelines. 

%%%%%%%%%%%%%%%%%%%%%%%%%%%%%%%%%%%%%%%%%%%%%%%%%%%%%%%%%%%%%%%%%%%%%%%%%%
\section*{Data Access Statement}
The imaging data were provided by the UK Biobank Resource under Application Number 17806. Researchers can apply to use the UK Biobank data resource for health-related research in the public interest. The  data set is freely available upon request at: \url{http://www.ukbiobank.ac.uk/}.

%%%%%%%%%%%%%%%%%%%%%%%%%%%%%%%%%%%%%%%%%%%%%%%%%%%%%%%%%%%%%%%%%%%%%%%%
% Acknowledgements
%%%%%%%%%%%%%%%%%%%%%%%%%%%%%%%%%%%%%%%%%%%%%%%%%%%%%%%%%%%%%%%%%%%%%%%%
\section*{Acknowledgements}
This work was supported by an EPSRC programme Grant (EP/P001009/1) and the Wellcome EPSRC Centre for Medical Engineering at the School of Biomedical Engineering and Imaging Sciences, King’s College London (WT 203148/Z/16/Z). This research has been conducted using the UK Biobank Resource under Application Number 17806. The GPU used in this research was generously donated by the NVIDIA Corporation.

%%%%%%%%%%%%%%%%%%%%%%%%%%%%%%%%%%%%%%%%%%%%%%%%%%%%%%%%%%%%%%%%%%%%%%%%
% References
%%%%%%%%%%%%%%%%%%%%%%%%%%%%%%%%%%%%%%%%%%%%%%%%%%%%%%%%%%%%%%%%%%%%%%%%
\bibliographystyle{model2-names} 
\section*{References}
%\bibliography{mybibfile,strings}
\bibliography{Media}

\end{document}